\documentclass[conference]{IEEEtran}
\IEEEoverridecommandlockouts
\usepackage{cite}
\usepackage{amsmath,amssymb,amsfonts}
\usepackage{algorithmic}
\usepackage{graphicx}
\usepackage{textcomp}
\usepackage{xcolor}
\usepackage{hyperref}
\usepackage{booktabs}
\usepackage{tabularx}

\def\BibTeX{{\rm B\kern-.05em{\sc i\kern-.025em b}\kern-.08em
    T\kern-.1667em\lower.7ex\hbox{E}\kern-.125emX}}

\begin{document}

\title{Brain Storm Optimization Based Swarm Learning for Diabetic Retinopathy Image
Classification}


\author{\IEEEauthorblockN{Liang Qu\textsuperscript{1}, Cunze Wang\textsuperscript{2}, and Yuhui Shi\textsuperscript{1}}
\IEEEauthorblockA{\textit{\textsuperscript{1}Department of Computer Science and Engineering, Southerun University of Science and Technology, Shenzhen, China} \\
\textit{\textsuperscript{2}School of Pharmacy, Fujian Medical University, Fuzhou, China}\\
qul@mail.sustech.edu.cn, ze1120792317@163.com, shiyh@sustech.edu.cn}
\thanks{Corresponding authors: Yuhui Shi (shiyh@sustech.edu.cn)}
}

\maketitle

\begin{abstract}
The application of deep learning techniques to medical problems has garnered widespread research interest in recent years, such as applying convolutional neural networks to medical image classification tasks. However, data in the medical field is often highly private, preventing different hospitals from sharing data to train an accurate model. Federated learning, as a privacy-preserving machine learning architecture, has shown promising performance in balancing data privacy and model utility by keeping private data on the client's side and using a central server to coordinate a set of clients for model training through aggregating their uploaded model parameters. Yet, this architecture heavily relies on a trusted third-party server, which is challenging to achieve in real life. Swarm learning, as a specialized decentralized federated learning architecture that does not require a central server, utilizes blockchain technology to enable direct parameter exchanges between clients. However, the mining of blocks requires significant computational resources, limiting its scalability. To address this issue, this paper integrates the brain storm optimization algorithm into the swarm learning framework, named BSO-SL. This approach clusters similar clients into different groups based on their model distributions.  Additionally, leveraging the architecture of BSO, clients are given the probability to engage in collaborative learning both within their cluster and with clients outside their cluster, preventing
the model from converging to local optima. The proposed method has been validated on a real-world diabetic retinopathy image classification dataset, and the experimental results demonstrate the effectiveness of the proposed approach.
\end{abstract}


\section{Introduction}

\begin{figure}
    \centering
    \includegraphics[width=1\linewidth]{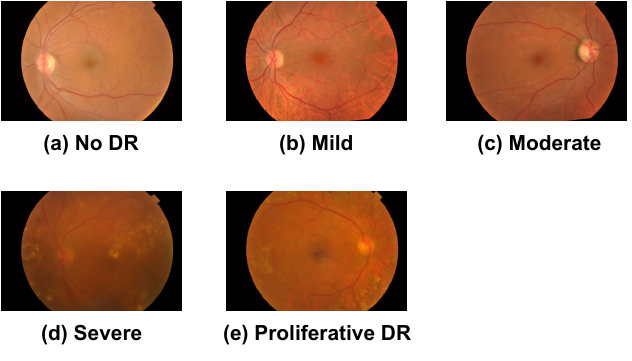}
    \caption{(a)-(e): different severities of diabetic retinopathy, ranging from No DR to Proliferative DR.}
    \label{fig:example}
\end{figure}

Diabetic retinopathy (DR) is a common disease caused by diabetes mellitus. Traditionally, doctors need to use fundus photography to capture patients' retina imagess, then classify the condition based on their experience and expertise, as shown in Figure \ref{fig:example} (a)-(e), a process that requires considerable manpower and relies heavily on the physician's knowledge \cite{abramoff2010automated}. In recent years, with the significant advancements in deep neural networks (DNNs) \cite{lecun2015deep,huang2021fapn,jiao2019survey}, particularly convolutional neural networks (CNNs) in image classification tasks, deep learning-based medical image classification methods \cite{shen2017deep,li2014medical,yadav2019deep,kandel2020transfer} have been widely applied to various medical imaging problems, thereby assisting doctors in rapid diagnosis. However, unlike traditional image classification problems where a vast amount of data can be labeled based on common sense, the availability of labeled medical imaging data is extremely limited in the field of medical image classification. This is due to the highly private nature of medical data, preventing hospitals from directly sharing patient medical data to train an accurate medical image classification model.

\begin{figure}
    \centering
    \includegraphics[width=1\linewidth]{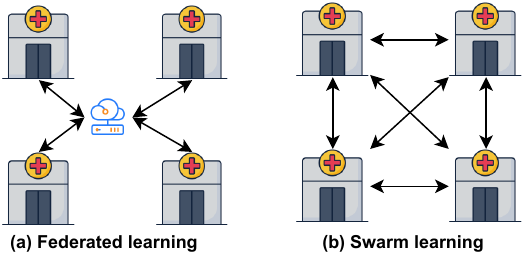}
    \caption{(a) the architecture of federated learning; (b) the architecture of swarm learning.}
    \label{fig:FLandSL}
\end{figure}

In recent years, federated learning \cite{li2020review,zhang2021survey} has shown significant advancement in the realm of privacy-centric machine learning tasks, notably within healthcare \cite{nguyen2022federated,wang2023fedmed} and recommendation systems \cite{yuan2024robust,qu2024personalized,yin2024ondevice,yuan2024hide,yuan2023hetefedrec,10.1145/3543507.3583337,zheng2024poisoning,zheng2024decentralized}.
By keeping private data on local clients and utilizing a central server to coordinate collaborative learning, federated learning allows for the sharing of model parameters through server-to-client communications. Specifically, as shown in Figure \ref{fig:FLandSL} (a), federated learning involves the following steps: (1) the server selects a group of clients to participate in the current round of model training and deploys the server's global parameters to the clients; (2) each client uses the parameters received from the server as the initial model parameters and trains the model based on their local data. They then update the model's parameters and upload them back to the server; (3) the server aggregates the parameters collected from the clients, using methods like the FedAvg \cite{mcmahan2017communication} aggregation algorithm, to update a global model, which is then redistributed to all clients. In this way, the privacy of each client's data is protected while indirectly leveraging the knowledge of other clients, thus addressing the issue of limited local data mentioned earlier. However, this federated learning architecture, which relies solely on server-client communications, depends heavily on a trusted central server, a requirement that is challenging to meet in real-world scenarios.

To address the limitations of traditional federated learning, swarm learning \cite{warnat2021swarm,saldanha2022swarm,warnat2020swarm}, a specialized decentralized federated learning architecture, has been proposed. As shown in Figure \ref{fig:FLandSL} (b), it introduces a blockchain-based client-to-client communication architecture for the model collaborative training, thereby reducing dependency on a central server. Specifically, clients strictly keep private data on their own clients for the local model training, and then the model is digitally signed and broadcasted to other clients. Subsequently, clients attempt to mine the current block upon receiving local models from others. Finally, with the receipt of the verified block, each clients updates its local model and prepares for the next round of communication. However, this blockchain-based swarm learning architecture demands significant computational resources for mining blocks, as well as considerable communication costs to interact with all other clients, thereby limiting its scalability.

Inspired by the collaborative characteristics of individuals in swarm intelligence algorithms \cite{yang2020swarm,kennedy2006swarm,eberhart2001swarm,he2020paired,he2020iterated,HE2021100894}, in this work, we integrate the Brain Storm Optimization (BSO) algorithm \cite{shi2011brain,cheng2016brain,cheng2019brain} into the swarm learning framework, termed BSO-SL. This integration employs clustering operations to assign similar clients to a group, enabling client-to-client collaborative learning within the group. This approach effectively addresses the scalability issues associated with blockchain-based swarm learning.
To facilitate this process, a server that is solely responsible for assigning neighbor clients but does not participate in model learning is introduced in this framework. This server clusters similar clients into the same group. Since the model's training parameters are private and high-dimensional, uploading them to the server poses risks of privacy breaches and incurs high communication costs due to the parameters' dimensionality. To overcome this, we assume that each model's parameter distribution follows a normal distribution. Therefore, only the distributional parameters, such as the mean and variance of each model's parameters, are uploaded to the server. This allows the server to cluster the clients based on the uploaded distributional information.
Moreover, the best-performing client in each group on the validation set is considered as the cluster's center. Utilizing the mechanism of the BSO algorithm, we exchange clients between different clusters with a certain probability to avoid local optima. 
Finally, within each group, clients can share their parameters with other group members, allowing standard parameter aggregation methods to be applied directly. 
In this way, BSO-SL not only maintains privacy and reduces computational and communication overhead but also enables effective collaboration among clients, enhancing the overall performance and scalability of the swarm learning framework.

Overall, the contributions of this work are summarized as below:
\begin{itemize}
    \item We propose the integration of Brain Storm Optimization (BSO) into the swarm learning framework, resulting in a novel approach named BSO-SL. This integration allows similar clients to be grouped together for direct client-to-client collaborative learning, enhancing the scalability of traditional blockchain-based swarm learning algorithms.
    \item Within the BSO-SL framework, we introduce a novel clustering strategy based on the distribution of clients' parameters. This strategy not only further protects client privacy but also improves communication efficiency.
    \item To validate the effectiveness of our proposed method, we conducted experiments on a real diabetic retinopathy image classification dataset. The experimental results demonstrate the effectiveness of our approach.
\end{itemize}

The remaining parts of this paper are organized as follows: Section 2 presents related work, offering insights into the existing literature and developments in this field. Section 3 provides a detailed description of our proposed method. Section 4 delves into our experimental setup and discusses the results obtained, followed by a conclusion in Section 5. 

\section{Related Work}

Swarm learning \cite{warnat2020swarm,warnat2021swarm,becker2022swarm}, as an emerging decentralized blockchain-based privacy-preserving machine learning technique (also regarded as a special form of federated learning characterized by exclusive client-to-client communications), has attracted considerable attention in recent years, particularly for its application in medical domains where privacy is a paramount concern. In this section, we will review some relevant methods of swarm learning in the medical field. For example, 
Warnat-Herresthal \textit{et al.} introduce a pioneering work in swarm learning \cite{warnat2021swarm}, leveraging edge computing and blockchain technology to collaboratively train multiple local models sourced from different hospitals in a decentralized manner for four heterogeneous diseases, resulting in eliminating the dependence on a central server. 
Saldanha \textit{et al.} \cite{saldanha2022swarm} propose to employ swarm learning to predict molecular alterations directly from histopathology slides. Saldanha \textit{et al.} \cite{saldanha2023direct} conduct a multicentric retrospective study using tissue samples from four patient cohorts from different countries, and utilize swarm learning for predicting molecular biomarkers in gastric cancer. 
Yuan \textit{et al.} \cite{yuan2023cooperative} introduces a swarm reinforcement learning to optimize the partitioning and offloading decisions in the DNN-empowered disease diagnostic process.
Yuan \textit{et al.} \cite{yuan2023cooperative} introduces a swarm reinforcement learning to optimize the partitioning and offloading decisions in the DNN-empowered disease diagnostic process.
However, most of current swarm learning methods are based on blockchain techniques, which demands significant computational resources for mining blocks, as well as considerable communication costs to interact with all other clients, thereby
limiting its scalability.


\begin{figure*}
    \centering
    \includegraphics[width=\linewidth]{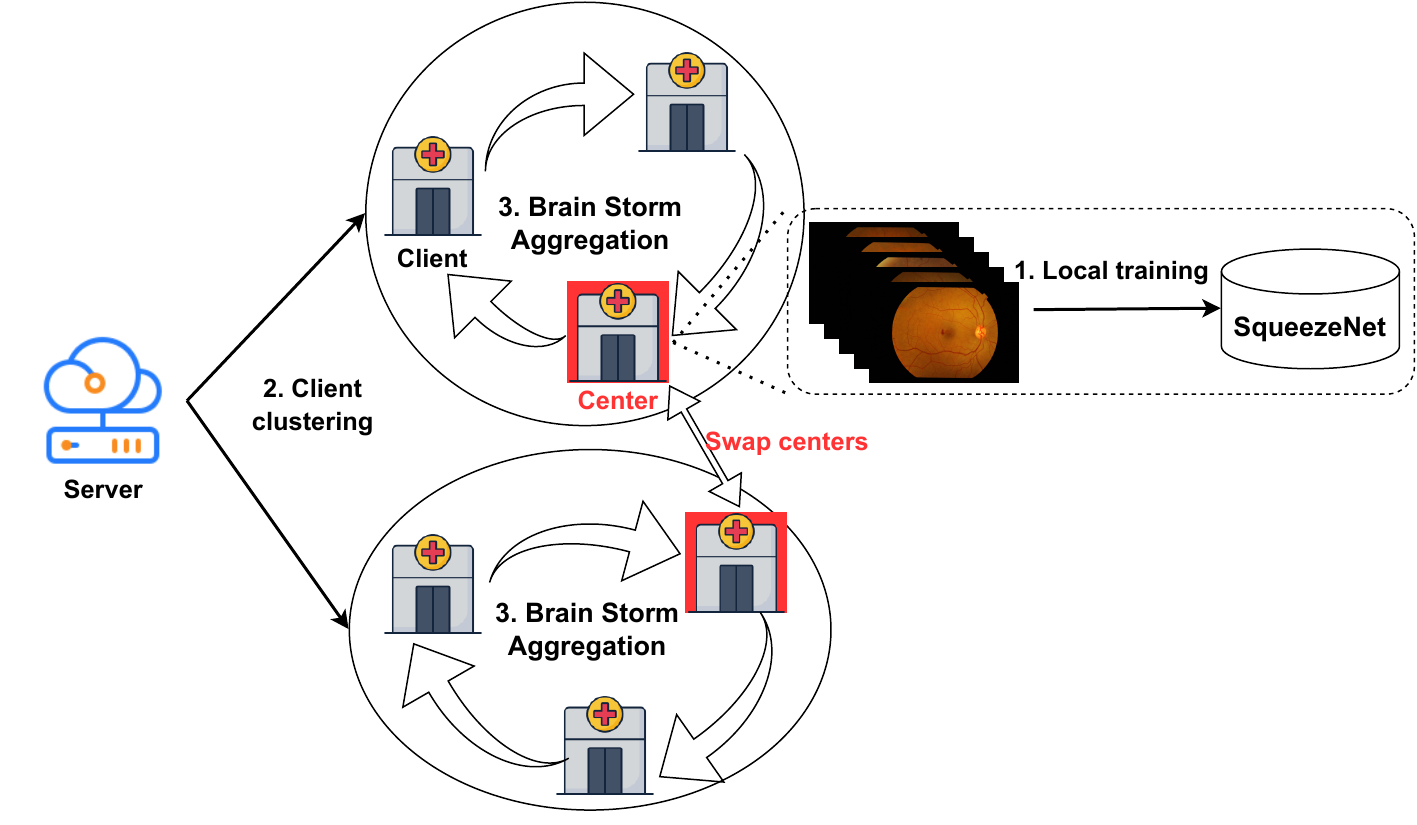}
    \caption{The architecture of the proposed BSO-SL.}
    \label{fig:BSO-SL}
\end{figure*}

\section{Proposed Method}
In this section, we will elaborate on our proposed BSO-SL algorithm. As illustrated in Figure \ref{fig:BSO-SL}, our algorithm primarily comprises three stages: (1) \textbf{Local training}: In this stage, each client conducts local model training using their private data. Upon completion of the training, clients upload the distribution of their model parameters to the server. (2) \textbf{Client clustering:} During this stage, the server assigns clients to a set of clusters based on the parameter distributions uploaded by each client. (3) \textbf{Brain storm aggregation:} This stage employs the framework of the Brain Storm Optimization algorithm. It involves exchanging individuals between different clusters with a certain probability and facilitating client-to-client communication to exchange parameter updates. This step is designed to prevent the algorithm from falling into local optima, thereby enhancing the effectiveness and robustness of the model updates.

\subsection{Local training}

Assuming $\mathcal{H}$ is a set of clients, each client $h \in \mathcal{H}$  possesses its own training dataset, denoted as $\mathcal{D}_{h} = \{\mathbf{X}_{h}, \mathbf{y}_{h}\}$, where $\mathbf{X}_{h} \in \mathbb{R}^{m \times n \times |\mathcal{D}_{h}|}$ and $\mathbf{y}_{h} \in \mathbb{R}^{|\mathcal{D}_{h}|}$ represent the training samples and their corresponding labels, respectively. Each training sample which is a medical image, has dimensions $m \times n$. Additionally, each client maintains a local model $f_{h}(\Theta_{h})$ parameterized by $\Theta_{h}$. During the local training phase, each client independently trains its model $f_{h}$ on its local dataset as below:
\begin{equation}
    \Theta_{h}^{*} = \mathop{\arg \min}\limits_{\Theta_{h}} \  \mathcal{L}_{u}(\mathcal{D}_{h},f_{u}(\Theta_{h}))
\end{equation}
where $\mathcal{L}_{u}$ is the loss function, e.g., the cross-entropy loss \cite{demirkaya2020exploring}.
Since our framework is model-agnostic, it can seamlessly integrate existing medical image classification neural network architectures. In this context, we choose the widely-used SqueezeNet model \cite{iandola2016squeezenet} as the local model $f_{h}(\Theta_{h})$ for each client.

\subsection{Client clustering}

Due to the limited data available under each client, it's challenging to train a highly accurate model. However, privacy concerns prevent direct data exchange between clients. To address this issue, the swarm learning architecture was proposed, allowing clients to retain their private data locally while exchanging model parameters or gradients via blockchain-based client-to-client communications. This method, however, has two main drawbacks: firstly, mining blocks for the blockchain requires extensive computational resources, and secondly, exchanging model parameters or gradients can still potentially leak client privacy \cite{zhu2019deep}.

To mitigate these issues, the approach of uploading only the distribution of local model parameters to the server for client clustering is proposed. It is assumed that the distribution of each local model's parameters approximates a Gaussian distribution \cite{liu2023federated}. Therefore, the mean and covariance of the model parameters are computed and uploaded. This method helps prevent the server from inferring detailed model parameters.

Furthermore, to resolve the scalability challenges inherent in the swarm learning, it is suggested that the server clusters clients based on the uploaded parameter distributions. Specifically, we employ the k-means algorithm \cite{ahmed2020k} to cluster the $|\mathcal{H}|$ clients into a set of clusters. Consequently, clients only need to interact with other clients within the same cluster, significantly improving the scalability of swarm learning. 

It's important to note that the role of the server in our BSO-SL architecture differs significantly from that in a traditional federated learning framework. In federated learning, the server plays a crucial role by aggregating parameters from all clients to learn a global model through server-to-client communications. In contrast, in the BSO-SL architecture, the server's function is limited to assigning neighbors. The collaborative learning of the model is entirely accomplished through client-to-client communications. This distinction is key because it enhances the scalability of the model. By reducing the reliance on a central server for aggregation and instead facilitating direct peer-to-peer interactions, BSO-SL can efficiently handle larger and more distributed networks of clients, making it more adaptable to scenarios with extensive and varied data sources.

\subsection{Brain storm aggregation}

After the client clustering, even though similar clients are grouped into the same clusters for swarm learning through client-to-client communications, which can address the issue of limited data for individual clients. However, the data across clients are often non-independent and identically distributed (non-IID). For example, the label distribution of data from each clinic may vary significantly, as demonstrated in Table \ref{tab:datasetsummary}, which presents the data distribution of  real-world Diabetic Retinopathy (DR) dataset \cite{choi2017multi} from different clinics.

From this table, we observe two key points: firstly, the number of training samples varies from one clinic to another, and secondly, the label distribution is different across clinics. For instance, in Clinic 1 (C1), the ``Moderate (2)'' category comprises the majority of samples, whereas in Clinic 14 (C14) and Clinic 4 (C4), there are no samples in the  ``Moderate (2)'' category.

This disparity poses a challenge: similar training data distributions can lead to similar model parameter distributions, potentially resulting in clinics with limited diversity in their data (such as Clinic 4 and Clinic 14) being clustered together. As a result, these clusters might not gain knowledge about the ``Moderate(2)'' category. This situation underscores the complexity of dealing with non-IID data in decentralized learning environments and highlights the need for strategies that can effectively manage the diversity of data distributions across different clients.

\begin{table*}[htbp]
  \large
  \centering
  \caption{Statistical Information of the DP Dataset: ``\#Samples'' represents the number of samples for each clinic's data. ``\#No DR (0)'', ``\#Mild (1)'', ``\#Moderate (2)'', ``\#Severe (3)'', and ``\#Proliferative DR (4)'' indicate the number of samples under each respective category. Clinics C1 to C14 represent different clinics.}
    \begin{tabular}{l|l|l|l|l|l|l|l|l|l|l|l|l|l|l}
    \toprule
          & C1 & C2 & C3 & C4 & C5 & C6 & C7 & C8 & C9 & C10 & C11 & C12 & C13 & C14 \\
    \midrule
    \midrule
    \#Samples & 410   & 638   & 974   & 351   & 141   & 533   & 287   & 92    & 61    & 52    & 42    & 34    & 28    & 14 \\
    \#NoDR(0) & 2     & 31    & 901   & 351   & 0     & 231   & 279   & 0     & 0     & 0     & 0     & 0     & 0     & 10 \\
    \#Mild(1) & 13    & 234   & 19    & 0     & 13    & 44    & 7     & 2     & 13    & 18    & 0     & 6     & 1     & 0 \\
    \#Moderate(2) & 307   & 233   & 39    & 0     & 91    & 165   & 1     & 63    & 28    & 11    & 33    & 3     & 22    & 0 \\
    \#Severe(3) & 32    & 60    & 2     & 0     & 6     & 47    & 0     & 9     & 1     & 4     & 5     & 21    & 3     & 2 \\
    \#ProliferativeDR(4) & 56    & 80    & 13    & 0     & 31    & 46    & 0     & 18    & 19    & 19    & 4     & 4     & 2     & 2 \\
    \bottomrule
    \end{tabular}%
  \label{tab:datasetsummary}%
\end{table*}%

To address this issue and drawing inspiration from swarm intelligence optimization algorithms, such as the Brain Storm Optimization (BSO) algorithm, where individuals from different clusters can still collaborate with a certain probability to prevent the algorithm from converging to local optima, we propose integrating the idea of BSO algorithm into our swarm learning framework for client parameter aggregation. This method is named Brain Storm Aggregation (BSA). Specifically, BSA consists of the following steps:
\begin{itemize}
    \item \textbf{Select Cluster Center:} After client clustering, clients within each cluster share their model performance on their local validation dataset, such as accuracy on the validation set. The best-performing client is selected as the center of each cluster.
    \item \textbf{Brain Storm:} As previously mentioned, allowing only similar users to collaborate might reduce the diversity in the swarm learning process, leading to missed knowledge and potential convergence to local optima. Here, we apply brain storm principles with two strategies to prevent this. First, for each cluster, we generate a random number $r_{1}$ and compare it with a predefined parameter $p_{1}$. If $r_{1} > p_{1}$, we randomly select a client within that cluster to replace the cluster center. Second, for the center of each cluster, we generate another random number $r_{2}$ and compare it with predefined parameter $p_{2}$. if $r_{2} > p_{2}$, we randomly select a center from another cluster and swap it with the current cluster's center. This ensures diversity in each cluster and prevents convergence to local optima.
    \item \textbf{Parameter Aggregation:} Once the clusters and their centers are determined, we can apply off-the-shelf parameter aggregation strategies, such as FedAvg \cite{mcmahan2017communication}, within each cluster to learn a global model $\Theta_{k}$ for the $k$-th cluster $\mathcal{G}_{k}$ via sharing model parameters among clients as below:
    \begin{equation}
        \Theta_{k}^{(t+1)} = \sum_{h \in \mathcal{G}_{k}}\frac{|\mathcal{D}_{h}|}{|\mathcal{D}_{\mathcal{G}_{k}}|}\Theta_{h}^{(t)}
    \end{equation}
    where $t$ represents the $t$-th training round, and $|\mathcal{D}_{\mathcal{G}_{k}}| = \sum_{h \in \mathcal{G}_{k}}|\mathcal{D}_{h}|$ is the total of training samples in the $k$-th cluster. After that, the aggregated global model parameter will be redistributed to each client within each group. Finally, each client uses the updated model parameters to initialize their local model for a new round of local training. This process is repeated until predefined stop criteria are met.
\end{itemize}

\section{Experiment}

In this section, to validate the effectiveness of our algorithm, we design experiments aimed at addressing the following research questions (RQs):
\begin{itemize}
    \item RQ1: How does our proposed method compare in performance with local training methods and federated learning approaches?
    \item RQ2: How do different local model architectures affect the performance of the model?
\end{itemize}

\subsection{Dataset}
To validate the effectiveness of the proposed method, we conduct experiments on a real-world Diabetic Retinopathy (DR) dataset \cite{choi2017multi}, which is available at \url{https://www.kaggle.com/c/aptos2019-blindness-detection/data}. Specifically, DR datasets consist of a large collection of retina images taken using fundus photography. Each image is classified by clinicians into one of five grades based on the severity of diabetic retinopathy, ranging from 0 to 4: No DR (0), Mild (1), Moderate (2), Severe (3), and Proliferative DR (4). Additionally, this dataset originates from multiple clinics, captured using different fundus photography equipment, resulting in varying retina image sizes. Since the original dataset does not specify the clinic source for each image, we assume that images of the same size originate from the same clinic. Moreover, we have filtered out sizes with fewer than 10 samples under the same size category. We randomly divided the dataset into three parts, allocating 80\% for the training set, 10\% for the validation set, and 10\% for the test set. Detailed information about the dataset is summarized in Table \ref{tab:datasetsummary}.

\subsection{Baselines}
In our experimental setup, we compare our method against three baseline approaches to provide a comprehensive evaluation:
\begin{itemize}
    \item \textbf{Centralized Method}: This method involves training a single centralized model using all available data, disregarding privacy concerns. It serves as a standard for assessing the maximum achievable performance when data is not subject to privacy constraints.
    \item \textbf{Local Training Method:} This method treats each client as an isolated individual, operating independently without any collaboration with other clients. It helps in assessing the effectiveness of models trained in complete isolation and sets a baseline for the improvement that collaborative approaches might offer.
    \item \textbf{Federated Learning Method:} This method implements the widely-used FedAvg algorithm, where collaborative learning is facilitated between clients and the server while each client's data remains local. 
\end{itemize}

\subsection{Setup}
For the local model of each client, we chose SqueezeNet \cite{iandola2016squeezenet}, widely used in image classification, due to its lightweight model size which effectively reduces communication costs. Additionally, since the image sizes from different clinics vary, each clinic resizes their images to a uniform dimension to align with the architecture of the SqueezeNet model. For the client clustering process, the k-means algorithm is selected as the clustering method. The number of clusters is set to 3. To generate the random numbers $r_{1}$ and $r_{2}$, we uniformly sample within the range of $[0,1]$. Moreover, the parameters $p_{1}$ and $p_{2}$ are set to 0.9 and 0.8, respectively. This setting is intended to balance exploration and exploitation during the aggregation process. 
Finally, the performance of our proposed algorithm is evaluated based on the average accuracy $acc$ achieved by each client on their respective local test sets as below:
\begin{equation}
    acc = \frac{1}{|\mathcal{H}|}\sum_{h \in \mathcal{H}}acc_{h}
\end{equation}
where $acc_{h}$ represents the classification accuracy of the $h$-th client on its local test set.

\subsection{Diabetic Retinopathy Image Classification (RQ1)}

We first discuss the performance of our model on the DR image classification task against other baseline methods. The experimental results are reported in Table \ref{tab:DRAcc}, from which we observe:
\begin{itemize}
    \item The centralized approach achieves the best results. This outcome is expected as the centralized method has access to all data for training, providing it with a substantial data advantage over other methods. However, it fails to consider data privacy issues, making it less feasible for real-world applications.
    \item The performance of the local training method is the poorest. This is due to the limited data available to each client, hindering the ability to train accurate models.
    \item Methods like FedAvg and our proposed approach perform better than local training but are inferior to the centralized method, which is a reasonable. It confirms that sharing model parameters can indirectly utilize the knowledge from other clients' data, striking a balance between privacy and model utility.
    \item Our method demonstrates competitive performance compared to FedAvg. This not only validates our approach's effectiveness in scenarios that do not rely on a central server but also enhances the scalability of the model. It underscores our method's potential in delivering promising results while maintaining decentralized architecture.
\end{itemize}

\begin{table}[htbp]
  \large
  \centering
  \caption{Performance Comparison of Different Methods on the DR Dataset with Respect to Accuracy Metric.}
    \begin{tabular}{lr}
    \toprule
    Method & \multicolumn{1}{l}{Acc} \\
    \midrule
    \midrule
    Centralized method & 0.4118 \\
    Local training method & 0.1924 \\
    FedAvg & 0.3719 \\
    BSO-SL & 0.3725 \\
    \bottomrule
    \end{tabular}%
  \label{tab:DRAcc}%
\end{table}%

\subsection{Model-agnostic analysis (RQ2)}
To validate the model-agnostic nature of our proposed framework, which can seamlessly integrate with existing image classification models, we conducted experiments using several different CNN architectures as the basis for local training models. These architectures include AlexNet \cite{krizhevsky2012imagenet}, VGG16 \cite{simonyan2014very}, InceptionV3 \cite{szegedy2016rethinking}, and the SqueezeNet model we initially employed. The experimental results are reported in Table \ref{tab:model-agnostic}, and from these results, we can observe:

\begin{itemize}
    \item The proposed BSO-SL architecture successfully integrates common CNN models, confirming its model-agnostic nature. This demonstrates the potential flexibility of our architecture to be applied to various image classification tasks across different domains.
    \item The architecture based on InceptionV3 achieved the best results among the tested models. This performance aligns with the results reported in other studies \cite{kandel2020transfer}. We attribute this high level of performance to the sophisticated and efficient design inherent in the InceptionV3 architecture.
    \item The SqueezeNet network, which we employed in our experiments, showed competitive performance compared to other architectures. A significant advantage of SqueezeNet is its lower number of network parameters. This aspect is particularly crucial in the context of swarm learning architectures where communication cost is a key consideration. The reduced parameter size of SqueezeNet makes it an efficient choice for scenarios where minimizing data transfer and maintaining model effectiveness are both priorities.
\end{itemize}

\begin{table}[htbp]
  \large
  \centering
  \caption{Performance Comparison of CNN Architectures.}
    \begin{tabular}{lr}
    \toprule
    Method & \multicolumn{1}{l}{Acc} \\
    \midrule
    \midrule
    BSO-SL(AlexNet) & 0.3703 \\
    BSO-SL(VGG16) & 0.4016 \\
    BSO-SL(InceptionV3) & 0.4216 \\
    BSO-SL(SqueezeNet) & 0.3725 \\
    \bottomrule
    \end{tabular}%
  \label{tab:model-agnostic}%
\end{table}%

\section{Conclusion}
To address the computational expense and scalability issues associated with blockchain-based swarm learning algorithms, we propose a Brain Storm Optimization-based Swarm Learning framework, BSO-SL. This framework clusters similar clients together based on their parameter distributions. Additionally, leveraging the architecture of BSO, clients are given the probability to engage in collaborative learning both within their cluster and with clients outside their cluster, preventing the model from converging to local optima. Finally, we validate our method on a real-world diabetic retinopathy image classification dataset. The experimental results demonstrate that our method achieves competitive performance compared to the FedAvg approach, while offering enhanced scalability due to its independence from a central server architecture.

In future work, we aim to extend the application of our method to datasets in other fields. 
Additionally, given the non-independent and identically distributed (non-IID) nature of each client's data and the heterogeneity in computational resources across clients, we aim to explore heterogeneous swarm learning in the future. This approach would involve clients adopting different model architectures to address the challenges posed by diverse data distributions.

\section{Acknowledgment}
This work is partially supported by the National Key R\&D Program of China under the Grant No. 2023YFE0106300 and 2017YFC0804002, Shenzhen Fundamental Research Program under the Grant No. JCYJ20200109141235597, and National Science Foundation of China under Grant No. 62250710682 and 61761136008.

\bibliographystyle{IEEEtran}
\bibliography{main}

\end{document}